\documentclass[10pt,twocolumn,letterpaper]{article}

 \usepackage{cvpr}              
\definecolor{cvprblue}{rgb}{0.21,0.49,0.74}
\usepackage[pagebackref,breaklinks,colorlinks,allcolors=cvprblue]{hyperref}
\usepackage{booktabs}
\usepackage[table]{xcolor}
\def\confName{CVPR}
\def\confYear{2027}

\title{MotionODE: Enhancing Autoregressive Text-to-Motion Generation with Continuous-Depth Modeling}

\author{
Zhaoxin Fan\\
Beihang University\\
{\tt\small zhaoxinf@buaa.edu.cn}
\and
Jian Yang\\
Beihang University
\and
Kejian Wu\\
XREAL
\and
Zeming Liu\\
Beihang University
\and
Wenjun Wu\\
Beihang University
\and
Lili Wang\\
Beihang University
}

\begin{document}
\twocolumn[{%
    \renewcommand\twocolumn[1][]{#1}%
    \maketitle

    \begin{center}
        \includegraphics[width=\textwidth]{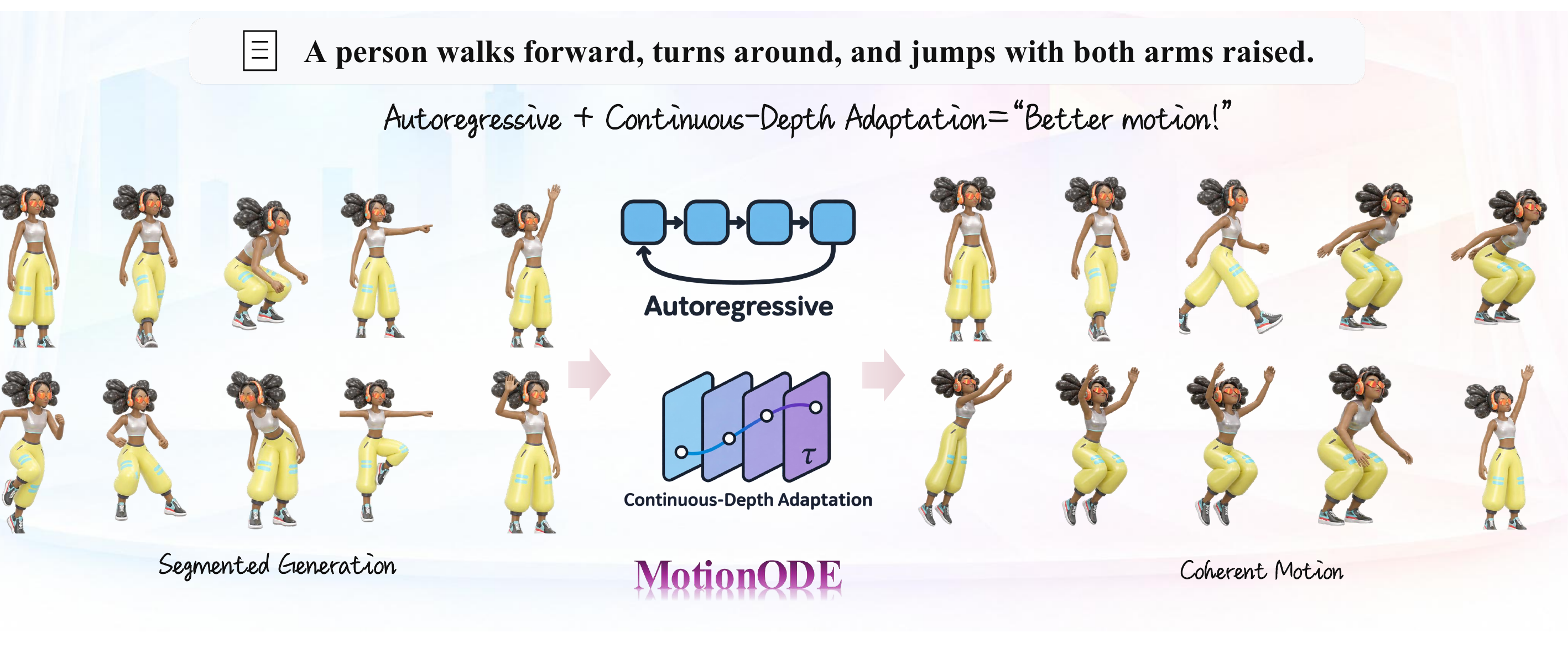}
        \vspace{-0.2in}
        \captionof{figure}{
            \textbf{MotionODE: continuous-depth adaptation for autoregressive text-to-motion generation.}
Given a text prompt describing multiple actions, MotionODE adapts the hidden-state evolution of pretrained autoregressive generators while preserving their original generation mechanisms.
The illustrated comparison highlights the transition from fragmented action sequences (left) to more coherent motion that follows the described action progression (right).
        }
        \label{fig:motionode_teaser}
    \end{center}
}]

\begin{abstract}
Autoregressive text-to-motion generation requires local motion predictions to reconcile cross-temporal textual requirements with the available motion context. While existing methods primarily improve motion representations and generation order, the internal formation of conditioned motion representations remains less explored. We introduce MotionODE, a continuous-depth post-training framework that refines hidden-state evolution in pretrained autoregressive generators. MotionODE reinterprets selected Transformer residual branches as local vector fields and replaces their original updates with midpoint Runge--Kutta (RK2) integration, while preserving the motion representation and outer generation mechanism. To address the mismatch between pretrained update paths and newly introduced intermediate states, we propose mixed one-step/two-step RK2 post-training under motion-token supervision, with optional distillation and update-map constraints to preserve pretrained behavior. This minimally invasive framework supports both next-token and next-scale generation without training the generator from scratch. Extensive experiments on HumanML3D demonstrate consistent improvements in text--motion alignment across MoScale, BAMM and LMR, highlighting internal representation updates as a complementary direction for improving autoregressive motion generation.
\end{abstract}    
\section{Introduction}
Text-to-motion generation aims to synthesize natural, semantically consistent human motion from language descriptions, with applications in character animation \cite{qing2023story}, human–computer interaction \cite{park2025unified}, and embodied intelligence \cite{xie2026textop}. It offers an intuitive way to control digital characters while reducing motion-authoring costs. The task requires translating action semantics and event ordering into coordinated movements, demanding both textual alignment and temporal coherence \cite{khani2025motion}.

Recent advances have established autoregressive generation over discrete motion representations as an effective approach to text-to-motion synthesis \cite{guo2024momask, zhang2024motiondiffuse, zou2024parco}. These methods encode continuous human motion into discrete tokens, predict motion tokens or token groups conditioned on text, and reconstruct the resulting sequence using a motion decoder. Existing approaches explore different generation mechanisms, including next-token \cite{jiang2023motiongpt}, grouped-token \cite{fu2026mogo}, and next-scale prediction \cite{zheng2026next}, while improving motion representations, conditioning mechanisms, and generation order. Nevertheless, the local prediction objective does not directly capture all requirements of the complete motion: an individual token or token group must contribute to semantic relationships that span multiple temporal segments. Accurate local predictions alone do not guarantee that the generated sequence respects the action order specified by the text or maintains coherent transitions between motion stages \cite{sui2026survey}. Coordinating global textual requirements with the evolving motion context thus remains an important challenge for autoregressive text-to-motion generation.

This challenge also depends on how conditioned motion representations are formed before each prediction: local token choices must reconcile cross-temporal textual requirements with the available motion context. While existing Transformers build these representations through discrete residual updates, their internal update trajectories offer an optimization dimension complementary to tokenization and generation order \cite{li2021ode}. We therefore ask: can redesigning hidden-state evolution improve text-conditioned motion generation while preserving the pretrained representation and autoregressive mechanism? 

To investigate this question, as shown in Fig. \ref{fig:motionode_teaser}, we introduce MotionODE,  a continuous-depth post-training framework for pretrained autoregressive text-to-motion generators, with two key components. First, MotionODE enables minimally invasive architectural adaptation. We reuse the pretrained weights of selected Transformer residual branches to define local vector fields, interpret their original residual updates as unit-step Euler-style updates, and introduce midpoint Runge–Kutta integration (RK2) to evolve the hidden states. This adaptation preserves the base model’s motion representation, text-conditioning interface, outer generation order, and motion decoder, without requiring a redesign or training the entire system from scratch. Because it operates on internal residual computation rather than a particular generation granularity, the framework is applicable to different autoregressive formulations, including next-token and next-scale generation. Second, we propose mixed one-step/two-step RK2 post-training. Simply replacing the integration scheme does not guarantee better generation: pretrained weights are optimized along the original discrete update path, whereas a new integrator queries different intermediate states and changes the inputs to subsequent branches. We therefore alternate between one-step and two-step RK2 over the same integration interval, training a shared set of parameters under ground-truth motion-token cross-entropy supervision. We further investigate prediction-distribution distillation and local update-map anchoring to balance adaptation to new integration paths with preservation of pretrained behavior. At inference, the generator uses a selected integration step count without requiring the original teacher model.

We evaluate MotionODE on HumanML3D using autoregressive generators with different generation mechanisms.  Extensive experiments demonstrate that MotionODE consistently improves text–motion alignment across existing autoregressive generators. Our main contributions are summarized as follows:

\begin{itemize}
    \item  We propose MotionODE, a continuous-depth post-training framework that improves pretrained autoregressive text-to-motion generators by reformulating selected residual updates as ODE integration, without changing their motion representations or generation mechanisms.
    \item We introduce mixed one-step/two-step RK2 post-training to adapt pretrained generators to new integration paths. Our minimally invasive design supports both next-token and next-scale generation.
    \item Extensive experiments on HumanML3D demonstrate consistent improvements in text–motion alignment across MoScale, BAMM and LMR.
\end{itemize}

\section{Related Work}
\label{sec:related}

\noindent\textbf{Text-to-motion generation.}
Existing methods primarily follow diffusion-based and autoregressive paradigms. Diffusion-based approaches progressively denoise motion under text conditioning: MotionDiffuse~\cite{zhang2024motiondiffuse} and MDM~\cite{tevet2022human} explore fine-grained conditioning and geometric supervision, while MLD~\cite{chen2023executing} improves efficiency through latent-space diffusion. Subsequent studies investigate accelerated sampling~\cite{dai2024motionlcm}, energy-based composition~\cite{zhang2025energymogen}, structured cross-modal interactions~\cite{hong2025salad}, hybrid diffusion--autoregressive modeling~\cite{meng2025rethinking}, and hierarchical flow matching~\cite{li2026motionhiflow}. Autoregressive methods instead formulate motion synthesis as discrete token prediction. TM2T~\cite{guo2022tm2t} introduces text--motion token translation, T2M-GPT~\cite{zhang2023generating} combines vector-quantized motion representations with a GPT-style generator, and MotionGPT~\cite{jiang2023motiongpt} unifies language and motion modeling. Later methods improve tokenization and generation through residual quantization and masked prediction~\cite{guo2024momask,pinyoanuntapong2024mmm}, model scaling~\cite{lu2025scamo}, hierarchical streaming generation~\cite{fu2026mogo}, and next-scale autoregression~\cite{zheng2026next}. Complementary approaches enrich generation with retrieved motion exemplars~\cite{zhang2023remodiffuse}, body-part-aware retrieval and fusion~\cite{kalakonda2025morag}, persistent temporal context~\cite{zhang2024infinimotion}, and retrieval-augmented masked modeling~\cite{li2026remomask}. Despite these advances in representations, generation mechanisms, and contextual information, the internal updates that form conditioned representations before autoregressive prediction remain less explored; MotionODE addresses this gap through minimally invasive continuous-depth adaptation and mixed RK2 post-training, while preserving the pretrained motion representation and generation mechanism.

\noindent\textbf{Neural ODE methods.}
Neural Ordinary Differential Equations~\cite{chen2018neural} model hidden-state evolution through continuous-depth dynamics parameterized by neural networks. Subsequent work improves their expressiveness through augmented state spaces~\cite{dupont2019augmented}, gradient accuracy through checkpointed differentiation~\cite{gholami2019anode}, and efficiency through dynamics regularization~\cite{finlay2020train}. ODE Transformer~\cite{li2021ode} extends this perspective to sequence modeling by introducing Runge--Kutta-inspired Transformer blocks. The relationship between discrete residual networks and continuous dynamics has also been studied through approximation analysis~\cite{sander2022residual} and implicit regularization during training~\cite{marion2024implicit}. More recently, Neural ODE Transformers~\cite{tong2025neural} parameterize attention and feed-forward weights as functions of continuous depth, enabling dynamical analysis and adaptive fine-tuning, while layer-parallel training~\cite{jiang2026layer} exploits an ODE formulation to parallelize Transformer training across depth. MotionODE shares their numerical-integration perspective on representation evolution, but reuses selected residual branches of pretrained autoregressive motion generators as local vector fields and adapts them through mixed one-step/two-step RK2 post-training, without introducing continuous-depth weight parameterization or changing the outer generation mechanism.
\begin{figure*}[t]
    \centering
    \includegraphics[width=\linewidth]{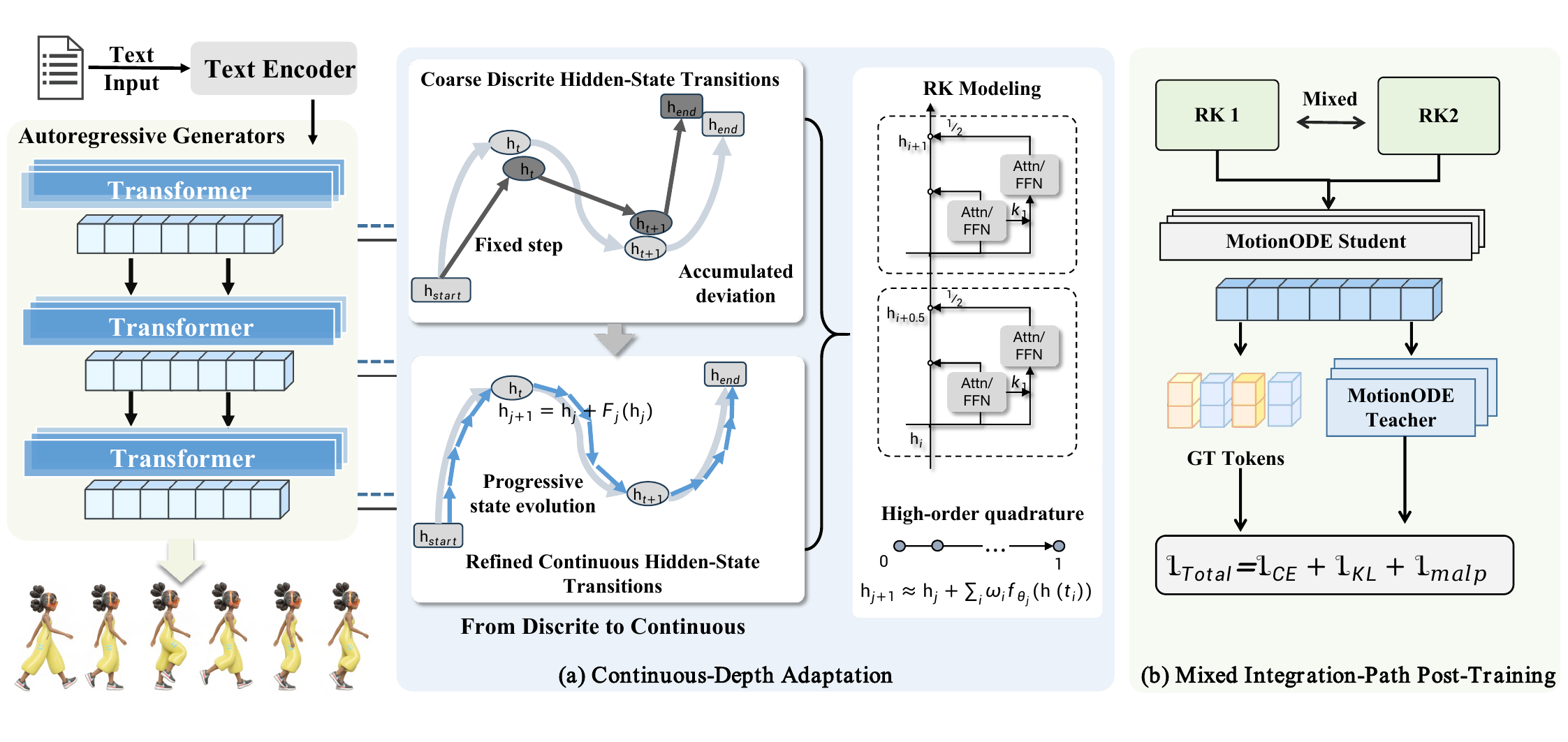}
    \vspace{-0.2in}
    \caption{\textbf{Overview of MotionODE.}
(a) \textbf{Continuous-depth adaptation.} Selected residual branches of a pretrained autoregressive motion generator are reinterpreted as local vector fields and integrated using midpoint RK2, refining internal hidden-state evolution while preserving the outer generation mechanism.
(b) \textbf{Mixed integration-path post-training.} A shared student alternates between one-step and two-step RK2 integration under ground-truth motion-token supervision. Optional prediction-distribution distillation and local update-map anchoring from a frozen pretrained teacher regularize deviations from the original model. At inference, the student uses a fixed integration step count without the teacher.}
    \label{fig:motionode_pipeline}
\end{figure*}

\section{Method}
\label{sec:method}

As shown in Fig. \ref{fig:motionode_pipeline}, we introduce MotionODE, a continuous-depth post-training framework for pretrained autoregressive text-to-motion generators. MotionODE reformulates selected Transformer residual updates as numerical integration and adapts the resulting model through mixed one-step/two-step RK2 training. The framework preserves the original motion tokenizer, text-conditioning interface, outer generation mechanism, and motion decoder.

\subsection{Background}
\label{sec:background}

\noindent\textbf{Autoregressive text-to-motion generation.}
Given a text description $c$, a motion tokenizer encodes a motion sequence $\mathbf{x}$ into discrete tokens or token groups $\mathbf{q}=(q_1,\ldots,q_K)$. An autoregressive generator models their conditional distribution as
\begin{equation}
    p_{\theta}(\mathbf{q}\mid c)
    =
    \prod_{k=1}^{K}
    p_{\theta}(q_k\mid q_{<k},c),
    \label{eq:ar_factorization}
\end{equation}
where $k$ indexes the outer generation stage. Depending on the backbone, $q_k$ can denote an individual motion token or a group of tokens at a particular scale. The generated tokens are subsequently reconstructed into continuous motion by the pretrained motion decoder.

Before each prediction, the generator forms a conditioned hidden representation through a sequence of Transformer residual updates. For a residual branch indexed by $b$, the update can be written as
\begin{equation}
    \mathbf{z}_{b+1}
    =
    \mathbf{z}_{b}
    +
    f_{\theta_b}(\mathbf{z}_{b};\mathcal{C}_k),
    \label{eq:residual_update}
\end{equation}
where $\mathcal{C}_k$ denotes the conditioning information available at generation stage $k$, including text, motion context, and attention constraints. The branch function $f_{\theta_b}$ includes the operations used by the original model, such as normalization, attention, or feed-forward transformation.

\noindent\textbf{Neural ordinary differential equations.}
Neural ODEs~\cite{chen2018neural} describe hidden-state evolution through a parameterized differential equation:
\begin{equation}
    \frac{\mathrm{d}\mathbf{z}(\tau)}{\mathrm{d}\tau}
    =
    f_{\theta}(\mathbf{z}(\tau),\tau;\mathcal{C}),
    \label{eq:neural_ode}
\end{equation}
where $\tau$ indexes network depth rather than physical motion time. A forward Euler step with step size $h$ gives
\begin{equation}
    \mathbf{z}(\tau+h)
    =
    \mathbf{z}(\tau)
    +
    h f_{\theta}(\mathbf{z}(\tau),\tau;\mathcal{C}).
\end{equation}
Thus, Eq.~\eqref{eq:residual_update} has the form of a unit-step Euler update. This interpretation motivates alternative integration schemes, but does not imply that pretrained residual networks already implement well-optimized continuous dynamics.

\subsection{Continuous-Depth Adaptation}
\label{sec:continuous_adaptation}

\noindent\textbf{Local vector fields from pretrained branches.}
Inspired by ODE Transformer~\cite{li2021ode}, we reinterpret selected residual branches as local vector fields. Unlike constructing a new continuous-depth generator, MotionODE directly reuses their pretrained parameters. Each selected branch defines an independent initial-value problem over a normalized interval $\tau\in[0,1]$:
\begin{equation}
    \frac{\mathrm{d}\mathbf{z}(\tau)}{\mathrm{d}\tau}
    =
    f_{\theta_b}(\mathbf{z}(\tau);\mathcal{C}_k),
    \qquad
    \mathbf{z}(0)=\mathbf{z}_b.
    \label{eq:local_ode}
\end{equation}
The terminal state becomes the input to the next branch:
$\mathbf{z}_{b+1}=\mathbf{z}(1)$.
Different branches retain their own parameters; parameter sharing occurs only across integration stages within the same branch. Unselected branches retain their original computation.

The conditioning information $\mathcal{C}_k$ is governed by the original backbone throughout integration. In particular, internal integration does not introduce additional motion predictions, alter attention masks, or change the outer autoregressive order. Existing text-conditioning modules and cache semantics are preserved.

\noindent\textbf{Midpoint RK2 integration.}
We approximate Eq.~\eqref{eq:local_ode} using the explicit midpoint Runge--Kutta method. For $S$ integration steps, we set $h=1/S$ and initialize $\mathbf{z}^{(0)}=\mathbf{z}_b$. Each step computes
\begin{equation}
\begin{aligned}
    \mathbf{k}_1^{(j)}
    &=
    f_{\theta_b}(\mathbf{z}^{(j)};\mathcal{C}_k),\\
    \mathbf{k}_2^{(j)}
    &=
    f_{\theta_b}
    \left(
        \mathbf{z}^{(j)}
        +\frac{h}{2}\mathbf{k}_1^{(j)};
        \mathcal{C}_k
    \right),\\
    \mathbf{z}^{(j+1)}
    &=
    \mathbf{z}^{(j)}+h\mathbf{k}_2^{(j)},
\end{aligned}
\label{eq:rk2}
\end{equation}
for $j=0,\ldots,S-1$. We denote the resulting branch update by
$\Phi_{\theta_b}^{(S)}(\mathbf{z}_b;\mathcal{C}_k)
=\mathbf{z}^{(S)}$.

Both $S=1$ and $S=2$ traverse the same interval $[0,1]$, but evaluate the vector field at different intermediate states. Without field correction, they require two and four branch evaluations, respectively. These evaluations occur entirely within the hidden-state computation preceding a motion prediction.

\noindent\textbf{Optional field correction.}
We additionally investigate a finite-difference correction to the inherited vector field. Define
\begin{equation}
    \mathcal{D}_{\epsilon}f(\mathbf{z})
    =
    \frac{
        f(\mathbf{z}+\epsilon f(\mathbf{z}))
        -
        f(\mathbf{z})
    }{\epsilon},
    \label{eq:directional_difference}
\end{equation}
which estimates the directional derivative of $f$ along its own update direction. The corrected field is
\begin{equation}
    g_{\theta_b}(\mathbf{z},\tau)
    =
    f_{\theta_b}(\mathbf{z})
    -
    \alpha\tau\,
    \mathcal{D}_{\epsilon}f_{\theta_b}(\mathbf{z}),
    \label{eq:field_correction}
\end{equation}
where conditioning arguments are omitted for clarity, $\alpha$ controls correction strength, and $\epsilon$ is the finite-difference displacement. When enabled, $g_{\theta_b}$ replaces $f_{\theta_b}$ in Eq.~\eqref{eq:rk2}, evaluated at the corresponding integration stage. This correction is an optional component rather than a prerequisite for continuous-depth adaptation, and its additional branch evaluations are included in computational comparisons.

\subsection{Mixed Integration-Path Post-Training}
\label{sec:post_training}

\noindent\textbf{Why post-training is necessary.}
Pretrained parameters are optimized for the original discrete residual trajectory. Replacing Euler-style updates with RK2 introduces intermediate states that were not encountered during pretraining and changes the inputs received by subsequent branches. Consequently, a higher-order integration scheme alone does not guarantee better motion generation. MotionODE addresses this mismatch by adapting a shared parameter set to multiple integration paths.

\noindent\textbf{Alternating one-step and two-step RK2.}
We initialize the student from the pretrained generator and alternate the integration step count across training microbatches:
\begin{equation}
    S_m =
    \begin{cases}
        1, & m \text{ is odd},\\
        2, & m \text{ is even},
    \end{cases}
    \label{eq:mixed_schedule}
\end{equation}
where $m$ is the microbatch index. Within each microbatch, all selected residual branches use the same step count. Both paths share the same parameters and integration interval, exposing the model to different intermediate states without introducing separate networks.

Using the original backbone's ground-truth conditioning procedure, we supervise the student with motion-token cross-entropy:
\begin{equation}
    \mathcal{L}_{\mathrm{CE}}^{(S)}
    =
    -\frac{1}{|\mathcal{V}|}
    \sum_{i\in\mathcal{V}}
    \log p_{\theta}^{(S)}
    \left(q_i^{*}\mid\mathcal{C}_i^{*}\right),
    \label{eq:ce_loss}
\end{equation}
where $\mathcal{V}$ contains valid supervised token positions, $q_i^{*}$ is the ground-truth token, and $\mathcal{C}_i^{*}$ is the conditioning context constructed by the original training pipeline. This formulation accommodates both next-token and next-scale supervision.

\noindent\textbf{Prediction-distribution distillation.}
To preserve pretrained predictive behavior, we optionally regularize the student using a frozen copy of the original generator. Let $p_0$ denote the teacher distribution computed along its original residual path. The distillation objective is
\begin{equation}
    \mathcal{L}_{\mathrm{KL}}^{(S)}
    =
    \frac{1}{|\mathcal{V}|}
    \sum_{i\in\mathcal{V}}
    D_{\mathrm{KL}}
    \left(
        p_0(\cdot\mid\mathcal{C}_i^{*})
        \,\Vert\,
        p_{\theta}^{(S)}(\cdot\mid\mathcal{C}_i^{*})
    \right).
    \label{eq:kl_loss}
\end{equation}
Cross-entropy remains the primary task supervision, while distillation constrains deviations from the pretrained output distribution.

\noindent\textbf{Local update-map anchoring.}
We further investigate a branch-level constraint that compares the student and teacher updates from the same incoming hidden state. For a selected branch $b$, the frozen teacher produces the original residual increment
\begin{equation}
    \mathbf{d}_b
    =
    f_{\theta_b^{0}}(\mathbf{z}_b;\mathcal{C}_k),
\end{equation}
while the student produces
$\mathbf{y}_b=\Phi_{\theta_b}^{(S)}(\mathbf{z}_b;\mathcal{C}_k)$.
We anchor the integrated student increment to the pretrained increment using
\begin{equation}
    \mathcal{L}_{\mathrm{map}}^{(S)}
    =
    \operatorname{Avg}_{\mathrm{valid}}
    \left[
        \frac{
            \operatorname{MSE}
            (\mathbf{y}_b-\mathbf{z}_b,\mathbf{d}_b)
        }{
            \operatorname{Mean}(\mathbf{d}_b^2)+\delta
        }
    \right],
    \label{eq:map_loss}
\end{equation}
where $\delta=10^{-6}$ ensures numerical stability. The average is taken over selected branches and valid token and sample positions, with feature dimensions reduced inside each term. Comparing both paths at the same input isolates changes in the local update map rather than conflating them with differences between complete network trajectories.

\begin{table*}[t]
    \centering
    \caption{\textbf{Quantitative comparison with autoregressive-related
    text-to-motion methods on HumanML3D.}
    The upper block presents reported results from prior work, while the lower
    block compares pretrained generators with their MotionODE counterparts
    under our evaluation settings.
    MotionODE uses one-step midpoint RK2 at inference.
    $\uparrow$ indicates higher is better, $\downarrow$ indicates lower is better,
    and $\rightarrow$ indicates closer to the real-motion value is better.
    Bold highlights the better result within each evaluated backbone pair.
    All values are rounded to three decimal places; -- denotes unavailable results.}
    \label{tab:main_results}
    \setlength{\tabcolsep}{4pt}
    \renewcommand{\arraystretch}{1.10}
    \small
    \begin{tabular*}{\textwidth}{@{\extracolsep{\fill}}lccccccc@{}}
        \toprule
        \textbf{Method}
        & \multicolumn{3}{c}{\textbf{R-Precision} $\uparrow$}
        & \textbf{FID} $\downarrow$
        & \textbf{MM-Dist} $\downarrow$
        & \textbf{Diversity} $\rightarrow$
        & \textbf{Multimodality} $\uparrow$ \\
        \cmidrule(lr){2-4}
        & \textbf{Top-1}
        & \textbf{Top-2}
        & \textbf{Top-3}
        & & & & \\
        \midrule

        Real Motion
        & 0.511 & 0.703 & 0.797
        & 0.002 & 2.974 & 9.503 & -- \\
        \midrule

        TM2T~\cite{guo2022tm2t}
        & 0.424 & 0.618 & 0.729
        & 1.501 & 3.467 & 8.589 & -- \\

        T2M-GPT~\cite{zhang2023generating}
        & 0.492 & 0.679 & 0.775
        & 0.141 & 3.121 & 9.722 & -- \\

        AttT2M~\cite{zhong2023attt2m}
        & 0.499 & 0.690 & 0.786
        & 0.112 & 3.038 & 9.700 & -- \\

        ParCo~\cite{zou2024parco}
        & 0.515 & 0.706 & 0.801
        & 0.109 & 2.927 & 9.576 & -- \\

        BAMM (Reported)~\cite{pinyoanuntapong2024bamm}
        & 0.525 & 0.720 & 0.814
        & 0.055 & 2.919 & 9.717 & -- \\
        \midrule

        LMR (Native) \cite{qian2026think}
        & 0.497
        & 0.687
        & 0.782
        & \textbf{0.039}
        & 3.098
        & \textbf{9.599}
        & 0.611 \\

        \rowcolor{blue!8}
        \textbf{LMR + MotionODE (Ours)}
        & \textbf{0.504}
        & \textbf{0.692}
        & \textbf{0.788}
        & 0.098
        & \textbf{3.066}
        & 9.748
        & 1.06754 \\
        \midrule

        BAMM (Native)
        & 0.522
        & 0.716
        & 0.811
        & 0.052
        & 2.939
        & 9.667
        & 1.782 \\

        \rowcolor{blue!8}
        \textbf{BAMM + MotionODE (Ours)}
        & \textbf{0.529}
        & \textbf{0.722}
        & \textbf{0.815}
        & \textbf{0.051}
        & \textbf{2.932}
        & \textbf{9.613}
        & \textbf{1.831} \\
        \midrule

        MoScale (Native)
        & 0.535
        & 0.726
        & 0.817
        & \textbf{0.048}
        & 2.836
        & \textbf{9.509}
        & 0.874 \\

        \rowcolor{blue!8}
        \textbf{MoScale + MotionODE (Ours)}
        & \textbf{0.548}
        & \textbf{0.735}
        & \textbf{0.826}
        & 0.092
        & \textbf{2.795}
        & 9.690
        & \textbf{0.909} \\

        \bottomrule
    \end{tabular*}
\end{table*}
\noindent\textbf{Training objective and inference.}
The complete post-training objective is
\begin{equation}
    \mathcal{L}
    =
    \mathcal{L}_{\mathrm{CE}}^{(S_m)}
    +
    \lambda_{\mathrm{KL}}\mathcal{L}_{\mathrm{KL}}^{(S_m)}
    +
    \lambda_{\mathrm{map}}\mathcal{L}_{\mathrm{map}}^{(S_m)},
    \label{eq:total_loss}
\end{equation}
where the auxiliary weights can be set to zero to recover pure mixed-RK2 training. We optimize through the explicit integration stages using standard backpropagation; teacher parameters remain frozen.

At inference, we select a fixed step count $S_{\mathrm{test}}$ and use it throughout generation. The teacher is discarded, and the original motion sampling and decoding pipeline is retained. Thus, MotionODE requires neither an inference-time teacher nor separate parameters for different integration step counts, although additional vector-field evaluations increase computational cost.


\section{Experiment}
\subsection{Experimental Details}
We evaluate MotionODE on HumanML3D, which represents human motion using
22 joints at 20 FPS.
We study MoScale, BAMM, LMR, following the standard training, validation,
and test splits distributed with the baselines.
Each baseline and its adapted counterpart share the same motion
representations, normalization statistics, evaluators, and data-loading
compatibility fixes.
Starting from the corresponding official pretrained checkpoints,
we update the motion-generation Transformers while freezing the motion
tokenizers and text encoders; BAMM's residual refiner is also frozen.
Selected self-attention and feed-forward residual branches are reformulated
using midpoint RK2 over a unit depth interval, with one and two integration
substeps alternated across microbatches.

Post-training uses AdamW with an effective batch size of 64,
a peak learning rate of $10^{-5}$, $\beta=(0.9,0.99)$,
weight decay of $10^{-5}$, and gradient clipping at 2.0.
We train for six epochs, comprising 2,232 updates, with a half-epoch warmup
followed by cosine decay to $10^{-6}$.
The training seed is fixed to 10711, and checkpoints are saved at epochs
2 and 6 for validation-based selection.
The full configuration combines the native cross-entropy objective
with frozen-teacher KL distillation and normalized Euler-map anchoring,
each weighted by 1, and uses a field-correction coefficient of 0.5
with a finite-difference step of 0.01; these additional components
are examined separately through ablations.
The teacher is removed at inference, and no inference parameters are added.
The selected MoScale, BAMM and LMR configurations use one RK2 substep with
CFG scales of 3, 5 and 1.5.
MoScale retains its native sampling schedule $[1,2,5,10]$.
We report R-Precision (Top-1/2/3), FID, matching distance, Diversity,
and Multimodality, averaging all 20 evaluation repetitions for each
fixed configuration.
\begin{figure*}[t]
    \centering
    \includegraphics[width=\linewidth]{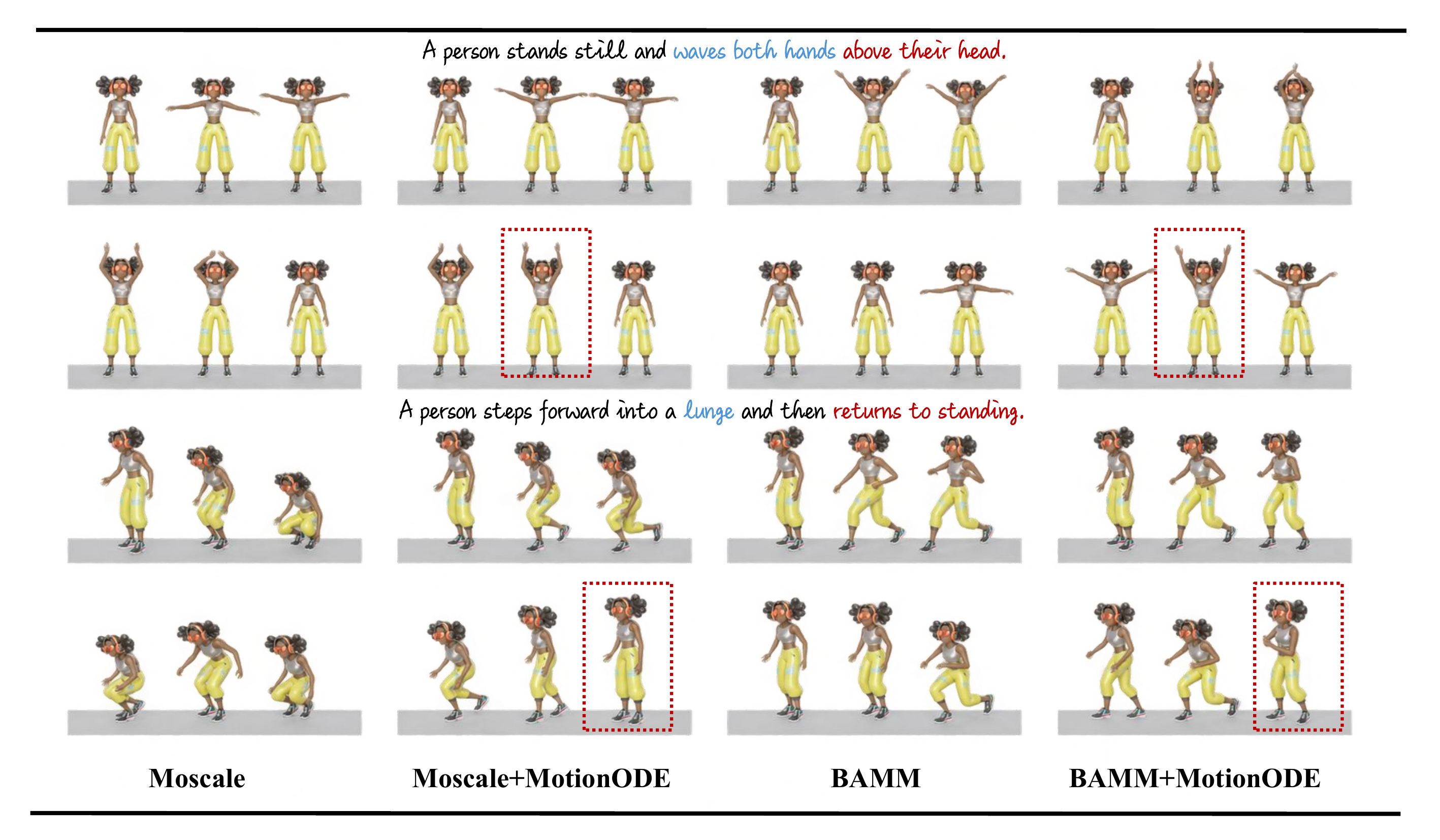
    }
    \vspace{-0.3in}
\caption{\textbf{Qualitative comparisons of autoregressive generators with and without MotionODE.} For both MoScale and BAMM, continuous-depth adaptation improves adherence to fine-grained action descriptions. Red dashed boxes highlight representative poses that reflect the specified
spatial constraint or final action state.}
\label{fig:qualitative_comparison}
\end{figure*}
\subsection{Main Result}
\noindent \textbf{Quantitative results.} Table~\ref{tab:main_results} evaluates MotionODE on three pretrained
backbones: LMR (Discrete), MoScale, and BAMM, with the latter two
representing next-scale and next-token motion generation, respectively.
Across all three backbones, MotionODE consistently improves
text-to-motion retrieval performance and reduces motion--text matching
distance, indicating better semantic alignment without changing the
original generation mechanism.
For LMR, R-Precision improves from 0.497 to 0.504,
from 0.687 to 0.692, and from 0.782 to 0.788 for Top-1, Top-2,
and Top-3, respectively, while matching distance decreases from
3.098 to 3.066.
These improvements extend the observed alignment benefits to an
additional pretrained generator.
The adapted MoScale and BAMM models also exhibit increased
multimodality, suggesting that their alignment gains do not come
at the expense of conditional motion diversity; this metric is
not yet available for the adapted LMR model.

These results are consistent with the motivation for combining
continuous-depth adaptation with integration-path-aware post-training:
midpoint RK2 evaluates residual branches at intermediate hidden states,
while mixed one-step/two-step training adapts pretrained parameters
to the resulting integration paths.
Rather than relying on a direct solver replacement, MotionODE jointly
adapts the hidden-state updates and their learned dynamics.
Importantly, the consistent alignment gains across the evaluated
backbones, including next-token and next-scale generation, suggest
that this approach is not tied to a particular autoregressive
factorization.
Nevertheless, distributional quality does not improve uniformly:
BAMM achieves a slightly lower FID, whereas both MoScale and
LMR (Discrete) exhibit an alignment--fidelity trade-off.
For LMR, FID increases from 0.039 to 0.098, and Diversity moves
farther from the real-motion reference, increasing from 9.599
to 9.748.
Thus, improved semantic alignment does not necessarily imply
better motion-distribution fidelity.
Overall, MotionODE provides a transferable strategy for improving
the semantic accuracy of pretrained autoregressive motion generators,
while highlighting the need to balance alignment and distributional
quality.

\begin{table*}[t]
    \centering
    \caption{\textbf{Ablation studies on MoScale.}
    Full denotes mixed one-step/two-step RK2 training with field correction,
    KL distillation, and map anchoring.
    All configurations are evaluated using CFG 3 and starting seed 2
    over 20 evaluation replications.
    Unless otherwise indicated, inference uses one-step RK2.
    Historical fixed-step models use their original training recipes.
    R-Precision is reported in percentages with four decimal places;
    other metrics are reported with five decimal places.
    $\uparrow$ denotes higher is better, $\downarrow$ denotes lower is better,
    and $\rightarrow$ denotes closer to the real-motion reference is better.}
    \label{tab:moscale_ablation}
    \setlength{\tabcolsep}{3pt}
    \renewcommand{\arraystretch}{1.08}
    \scriptsize
    \begin{tabular*}{\textwidth}{@{\extracolsep{\fill}}lccccccc@{}}
        \toprule
        \textbf{Configuration}
        & \multicolumn{3}{c}{\textbf{R-Precision (\%)} $\uparrow$}
        & \textbf{FID} $\downarrow$
        & \textbf{MM-Dist} $\downarrow$
        & \textbf{Diversity} $\rightarrow$
        & \textbf{Multimodality} $\uparrow$ \\
        \cmidrule(lr){2-4}
        & \textbf{Top-1}
        & \textbf{Top-2}
        & \textbf{Top-3}
        & & & & \\
        \midrule

        \rowcolor{blue!8}
        \textbf{MotionODE (Full)}
        & 54.7551 & 73.5428 & 82.6031
        & 0.09190 & 2.79532 & 9.68971 & 0.90905 \\

        \midrule
        \multicolumn{8}{l}{\textit{(a) Auxiliary losses}} \\

        w/o map anchor
        & 54.7196 & 73.6115 & 82.5177
        & 0.09182 & 2.79524 & 9.67314 & 0.91561 \\

        w/o KL
        & 54.5623 & 73.6625 & 82.5299
        & 0.11190 & 2.79473 & 9.69401 & 0.89622 \\

        w/o KL and map anchor
        & 54.3761 & 73.5051 & 82.5809
        & 0.11146 & 2.79975 & 9.68555 & 0.90912 \\

        \midrule
        \multicolumn{8}{l}{\textit{(b) Field correction and component interactions}} \\

        w/o field correction
        & 54.7839 & 73.4774 & 82.4856
        & 0.08908 & 2.78883 & 9.69412 & 0.91850 \\

        w/o field correction and map anchor
        & 54.8305 & 73.7223 & 82.6906
        & 0.08829 & 2.78423 & 9.68761 & 0.91099 \\

        w/o field correction and KL
        & 54.5357 & 73.6226 & 82.6108
        & 0.10994 & 2.79307 & 9.68528 & 0.89485 \\

        w/o field correction, KL, and map anchor
        & 54.6797 & 73.4863 & 82.5465
        & 0.10819 & 2.79308 & 9.69130 & 0.89415 \\

        \midrule
        \multicolumn{8}{l}{\textit{(c) Matched student field-evaluation budget}} \\

        Budget-matched control / Two-step RK2
        & 54.6576 & 73.5749 & 82.5576
        & 0.08827 & 2.78792 & 9.69110 & 0.90286 \\

        \midrule
        \multicolumn{8}{l}{\textit{(d) Historical training-step configurations}} \\

        Fixed one-step training / One-step RK2
        & 54.7861 & 73.7356 & 82.6961
        & 0.11300 & 2.78318 & 9.69659 & 0.88848 \\

        Fixed two-step training / Two-step RK2
        & 54.9047 & 73.9827 & 82.7870
        & 0.11241 & 2.77656 & 9.71332 & 0.82794 \\

        \bottomrule
    \end{tabular*}
\end{table*}
\noindent\textbf{Qualitative results.}
Fig.~\ref{fig:qualitative_comparison} illustrates how MotionODE improves
the semantic fidelity of existing autoregressive generators.
For the hand-waving prompt, the original models only partially realize
the specified motion: MoScale frequently places the arms near shoulder
height, while BAMM does not consistently sustain the overhead gesture
across the displayed poses.
After adaptation, both models exhibit clearer overhead hand movements,
better reflecting the spatial constraint ``above their head.''
For the lunge prompt, the original MoScale produces a predominantly
squat-like sequence, and BAMM does not clearly complete the requested
return to standing.
Their adapted counterparts better express the forward-lunge posture
and conclude in an upright stance.
These examples suggest that continuous-depth adaptation helps pretrained
autoregressive generators capture both fine-grained pose requirements
and the progression of multi-stage actions.
Importantly, these improvements arise from adapting hidden-state evolution,
without replacing the underlying tokenization or autoregressive
generation mechanism.


\subsection{Ablation Study}
\label{sec:ablation}

We conduct ablation studies on MoScale to examine auxiliary losses,
field correction, and integration-step configurations.
Table~\ref{tab:moscale_ablation} reports results averaged over
20 evaluation replications with CFG 3 and starting seed 2.
We use one-step RK2 inference unless otherwise specified.

\noindent\textbf{Auxiliary losses.}
KL distillation is the most consistently beneficial auxiliary constraint.
Across four pairs with the field-correction and map-anchoring settings
held fixed, retaining KL improves Top-1 retrieval and reduces FID.
In the full configuration, removing KL noticeably degrades FID and
also reduces multimodality.
These results are consistent with KL providing an output-level constraint
that limits distributional drift during integration-path adaptation.
Map anchoring has a smaller and more configuration-dependent effect.
With field correction enabled, retaining it improves Top-1 and Top-3,
while FID and matching distance remain nearly unchanged.
Without KL, map anchoring also recovers part of the Top-1 degradation,
although it does not improve every metric.
Thus, the two constraints play different roles:
KL provides the more consistent alignment--fidelity benefit,
whereas map anchoring offers a local regularization effect whose utility
depends on the remaining components.

\noindent\textbf{Field correction and interactions.}
The field-correction ablations reveal a trade-off rather than a uniform gain.
Relative to removing field correction alone, the full configuration
achieves better Top-2 and Top-3 retrieval and a diversity value closer
to the real-motion reference, whereas the uncorrected variant obtains
slightly better Top-1, FID, matching distance, and multimodality.
The interaction with map anchoring is particularly important:
removing both components improves retrieval and FID relative to the
full configuration.
This indicates that field correction and local update anchoring should
not be interpreted as independently additive improvements.
Meanwhile, KL remains beneficial with or without these components,
confirming that its contribution is more robust to the choice of
hidden-state adaptation mechanism.

\noindent\textbf{Matched field-evaluation budget.}
Under the matched student field-evaluation budget, the full configuration
with one-step RK2 obtains better Top-1 and Top-3 retrieval and higher
multimodality than the uncorrected two-step RK2 control.
The two-step control instead achieves lower FID and matching distance.
Consequently, spending the available branch evaluations on additional
integration substeps is not uniformly preferable to the corrected
one-step configuration.
The comparison exposes two distinct uses of the evaluation budget:
the selected configuration favors retrieval accuracy and conditional
variation, while the two-step alternative favors distributional fidelity.
Here, the matched budget refers to student field evaluations rather
than measured end-to-end runtime.

\noindent\textbf{Training-step configurations.}
Compared with the historical fixed-step checkpoints, mixed-step training
followed by one-step RK2 inference yields lower FID, higher multimodality,
and diversity closer to the real-motion reference.
The fixed-step configurations achieve somewhat stronger retrieval
and lower matching distance, but these gains are accompanied by reduced
conditional variation, particularly for the two-step checkpoint.
The results therefore favor the mixed-step configuration when
distributional fidelity and conditional diversity are considered
alongside semantic alignment.
Since the historical checkpoints use different training recipes,
this comparison characterizes the resulting configurations rather than
isolating the training-step schedule alone.

\noindent\textbf{Final configuration.}
We adopt mixed one-step/two-step RK2 training and one-step RK2 inference
as our final operating configuration.
Among the evaluated training-step configurations, it provides a favorable
balance between text--motion alignment, distributional fidelity, and
conditional diversity, without requiring two integration steps at inference.
The budget-matched comparison further shows that its retrieval and
multimodality advantages are not obtained simply by assigning more
student field evaluations.
Nevertheless, the component ablations demonstrate that the full
configuration is not uniformly optimal across all quality metrics.
We therefore present it as a selected quality--computation trade-off,
rather than claiming dominance over every tested variant.

\section{Conclusion}
\label{sec:conclusion}

We have presented MotionODE, a continuous-depth post-training framework
for pretrained autoregressive text-to-motion generators.
Inspired by ODE Transformer, MotionODE reinterprets selected residual
branches as local vector fields and adapts their hidden-state evolution
through mixed one-step/two-step midpoint RK2 training, while preserving
the original tokenization, conditioning, and generation mechanisms.
Experiments on LMR (Discrete), MoScale, and BAMM show improved
text--motion alignment across multiple backbones, including next-scale
and next-token generation, supporting the applicability of this
framework to different autoregressive factorizations.
Qualitative comparisons further illustrate improved adherence to
fine-grained pose constraints and the progression of multi-stage actions.
Together, these findings highlight integration-path-aware post-training
as a practical approach to improving the semantic accuracy of existing
motion generators without redesigning their generation pipelines.

\noindent\textbf{Limitations.}
Our evaluation focuses on HumanML3D, leaving generalization to broader
motion domains and more diverse textual descriptions for future study.
Moreover, alignment improvements do not uniformly translate into lower
FID, motivating further investigation of the balance between semantic
accuracy and motion-distribution fidelity.


{
    \small
    \bibliographystyle{ieeenat_fullname}
    \bibliography{main}

\begin{thebibliography}{38}
\providecommand{\natexlab}[1]{#1}
\providecommand{\url}[1]{\texttt{#1}}
\expandafter\ifx\csname urlstyle\endcsname\relax
  \providecommand{\doi}[1]{doi: #1}\else
  \providecommand{\doi}{doi: \begingroup \urlstyle{rm}\Url}\fi

\bibitem[Chen et~al.(2018)Chen, Rubanova, Bettencourt, and Duvenaud]{chen2018neural}
Ricky~TQ Chen, Yulia Rubanova, Jesse Bettencourt, and David~K Duvenaud.
\newblock Neural ordinary differential equations.
\newblock \emph{Advances in neural information processing systems}, 31, 2018.

\bibitem[Chen et~al.(2023)Chen, Jiang, Liu, Huang, Fu, Chen, and Yu]{chen2023executing}
Xin Chen, Biao Jiang, Wen Liu, Zilong Huang, Bin Fu, Tao Chen, and Gang Yu.
\newblock Executing your commands via motion diffusion in latent space.
\newblock In \emph{2023 IEEE/CVF Conference on Computer Vision and Pattern Recognition (CVPR)}, pages 18000--18010. IEEE, 2023.

\bibitem[Dai et~al.(2024)Dai, Chen, Wang, Liu, Dai, and Tang]{dai2024motionlcm}
Wenxun Dai, Ling-Hao Chen, Jingbo Wang, Jinpeng Liu, Bo Dai, and Yansong Tang.
\newblock Motionlcm: Real-time controllable motion generation via latent consistency model.
\newblock In \emph{European Conference on Computer Vision}, pages 390--408. Springer, 2024.

\bibitem[Dupont et~al.(2019)Dupont, Doucet, and Teh]{dupont2019augmented}
Emilien Dupont, Arnaud Doucet, and Yee~Whye Teh.
\newblock Augmented neural odes.
\newblock \emph{Advances in neural information processing systems}, 32, 2019.

\bibitem[Finlay et~al.(2020)Finlay, Jacobsen, Nurbekyan, and Oberman]{finlay2020train}
Chris Finlay, J{\"o}rn-Henrik Jacobsen, Levon Nurbekyan, and Adam Oberman.
\newblock How to train your neural ode: the world of jacobian and kinetic regularization.
\newblock In \emph{International conference on machine learning}, pages 3154--3164. PMLR, 2020.

\bibitem[Fu et~al.(2026)Fu, Sun, Fang, Cai, and Kim]{fu2026mogo}
Dongjie Fu, Tengjiao Sun, Pengcheng Fang, Xiaohao Cai, and Hansung Kim.
\newblock Mogo: Residual quantized hierarchical causal transformer for real-time and infinite-length 3d human motion generation.
\newblock In \emph{Proceedings of the AAAI Conference on Artificial Intelligence}, pages 3994--4002, 2026.

\bibitem[Gholami et~al.(2019)Gholami, Keutzer, and Biros]{gholami2019anode}
Amir Gholami, Kurt Keutzer, and George Biros.
\newblock Anode: Unconditionally accurate memory-efficient gradients for neural odes.
\newblock \emph{arXiv preprint arXiv:1902.10298}, 2019.

\bibitem[Guo et~al.(2022)Guo, Zuo, Wang, and Cheng]{guo2022tm2t}
Chuan Guo, Xinxin Zuo, Sen Wang, and Li Cheng.
\newblock Tm2t: Stochastic and tokenized modeling for the reciprocal generation of 3d human motions and texts.
\newblock In \emph{European conference on computer vision}, pages 580--597. Springer, 2022.

\bibitem[Guo et~al.(2024)Guo, Mu, Javed, Wang, and Cheng]{guo2024momask}
Chuan Guo, Yuxuan Mu, Muhammad~Gohar Javed, Sen Wang, and Li Cheng.
\newblock Momask: Generative masked modeling of 3d human motions.
\newblock In \emph{2024 IEEE/CVF Conference on Computer Vision and Pattern Recognition (CVPR)}, pages 1900--1910. IEEE, 2024.

\bibitem[Hong et~al.(2025)Hong, Kim, Yoon, Nam, Cha, and Noh]{hong2025salad}
Seokhyeon Hong, Chaelin Kim, Serin Yoon, Junghyun Nam, Sihun Cha, and Junyong Noh.
\newblock Salad: Skeleton-aware latent diffusion for text-driven motion generation and editing.
\newblock In \emph{2025 IEEE/CVF Conference on Computer Vision and Pattern Recognition (CVPR)}, pages 7158--7168. IEEE, 2025.

\bibitem[Jiang et~al.(2023)Jiang, Chen, Liu, Yu, Yu, and Chen]{jiang2023motiongpt}
Biao Jiang, Xin Chen, Wen Liu, Jingyi Yu, Gang Yu, and Tao Chen.
\newblock Motiongpt: Human motion as a foreign language.
\newblock \emph{Advances in Neural Information Processing Systems}, 36:\penalty0 20067--20079, 2023.

\bibitem[Jiang et~al.(2026)Jiang, Salvad{\'o}-Benasco, Cyr, Kopani{\v{c}}{\'a}kov{\'a}, Krause, and Schroder]{jiang2026layer}
Shuai Jiang, Marc Salvad{\'o}-Benasco, Eric~C Cyr, Alena Kopani{\v{c}}{\'a}kov{\'a}, Rolf Krause, and Jacob~B Schroder.
\newblock Layer-parallel training for transformers.
\newblock \emph{arXiv preprint arXiv:2601.09026}, 2026.

\bibitem[Kalakonda et~al.(2025)Kalakonda, Maheshwari, and Sarvadevabhatla]{kalakonda2025morag}
Sai~Shashank Kalakonda, Shubh Maheshwari, and Ravi~Kiran Sarvadevabhatla.
\newblock Morag-multi-fusion retrieval augmented generation for human motion.
\newblock In \emph{2025 IEEE/CVF Winter Conference on Applications of Computer Vision (WACV)}, pages 4564--4573. IEEE, 2025.

\bibitem[Khani et~al.(2025)Khani, Rampini, Roy, Nadela, Kaplan, Atherton, Cheung, and Bibliowicz]{khani2025motion}
Aliasghar Khani, Arianna Rampini, Bruno Roy, Larasika Nadela, Noa Kaplan, Evan Atherton, Derek Cheung, and Jacky Bibliowicz.
\newblock Motion generation: A survey of generative approaches and benchmarks.
\newblock \emph{arXiv preprint arXiv:2507.05419}, 2025.

\bibitem[Li et~al.(2021)Li, Du, Zhou, Zhou, Zeng, Xiao, and Zhu]{li2021ode}
Bei Li, Quan Du, Tao Zhou, Shuhan Zhou, Xin Zeng, Tong Xiao, and Jingbo Zhu.
\newblock Ode transformer: An ordinary differential equation-inspired model for neural machine translation.
\newblock \emph{arXiv preprint arXiv:2104.02308}, 2021.

\bibitem[Li et~al.(2026{\natexlab{a}})Li, Lin, Zeng, Kang, Li, and Hu]{li2026motionhiflow}
Heng Li, Xiaotong Lin, Ling-An Zeng, Yulei Kang, Shuai Li, and Jian-Fang Hu.
\newblock Motionhiflow: Text-to-motion via hierarchical flow matching.
\newblock \emph{arXiv preprint arXiv:2604.23264}, 2026{\natexlab{a}}.

\bibitem[Li et~al.(2026{\natexlab{b}})Li, Wang, Zhang, and Tang]{li2026remomask}
Zhengdao Li, Siheng Wang, Zeyu Zhang, and Hao Tang.
\newblock Remomask: Retrieval-augmented masked motion generation.
\newblock In \emph{European Conference on Computer Vision}, pages 111--128. Springer, 2026{\natexlab{b}}.

\bibitem[Lu et~al.(2025)Lu, Wang, Lu, Chen, Dai, Dong, Dou, Dai, and Zhang]{lu2025scamo}
Shunlin Lu, Jingbo Wang, Zeyu Lu, Ling-Hao Chen, Wenxun Dai, Junting Dong, Zhiyang Dou, Bo Dai, and Ruimao Zhang.
\newblock Scamo: Exploring the scaling law in autoregressive motion generation model.
\newblock In \emph{2025 IEEE/CVF Conference on Computer Vision and Pattern Recognition (CVPR)}, pages 27872--27882. IEEE, 2025.

\bibitem[Marion et~al.(2024)Marion, Wu, Sander, and Biau]{marion2024implicit}
Pierre Marion, Yu-Han Wu, Michael Sander, and G{\'e}rard Biau.
\newblock Implicit regularization of deep residual networks towards neural odes.
\newblock In \emph{International Conference on Learning Representations}, pages 40795--40834, 2024.

\bibitem[Meng et~al.(2025)Meng, Xie, Peng, Han, and Jiang]{meng2025rethinking}
Zichong Meng, Yiming Xie, Xiaogang Peng, Zeyu Han, and Huaizu Jiang.
\newblock Rethinking diffusion for text-driven human motion generation: Redundant representations, evaluation, and masked autoregression.
\newblock In \emph{2025 IEEE/CVF Conference on Computer Vision and Pattern Recognition (CVPR)}, pages 27859--27871. IEEE, 2025.

\bibitem[Park et~al.(2025)Park, Choi, and Yun]{park2025unified}
Jeongeun Park, Sungjoon Choi, and Sangdoo Yun.
\newblock A unified framework for motion reasoning and generation in human interaction.
\newblock In \emph{2025 IEEE/CVF International Conference on Computer Vision (ICCV)}, pages 10698--10707. IEEE, 2025.

\bibitem[Pinyoanuntapong et~al.(2024{\natexlab{a}})Pinyoanuntapong, Saleem, Wang, Lee, Das, and Chen]{pinyoanuntapong2024bamm}
Ekkasit Pinyoanuntapong, Muhammad~Usama Saleem, Pu Wang, Minwoo Lee, Srijan Das, and Chen Chen.
\newblock Bamm: Bidirectional autoregressive motion model.
\newblock In \emph{European Conference on Computer Vision}, pages 172--190. Springer, 2024{\natexlab{a}}.

\bibitem[Pinyoanuntapong et~al.(2024{\natexlab{b}})Pinyoanuntapong, Wang, Lee, and Chen]{pinyoanuntapong2024mmm}
Ekkasit Pinyoanuntapong, Pu Wang, Minwoo Lee, and Chen Chen.
\newblock Mmm: Generative masked motion model.
\newblock In \emph{2024 IEEE/CVF Conference on Computer Vision and Pattern Recognition (CVPR)}, pages 1546--1555. IEEE, 2024{\natexlab{b}}.

\bibitem[Qian et~al.(2026)Qian, Wang, Feng, Xu, Lu, Liu, Sun, Chen, Liu, and Wang]{qian2026think}
Yijie Qian, Juncheng Wang, Yuxiang Feng, Chao Xu, Wang Lu, Yang Liu, Baigui Sun, Yiqiang Chen, Yong Liu, and Shujun Wang.
\newblock Think before you move: Latent motion reasoning for text-to-motion generation.
\newblock \emph{IEEE transactions on pattern analysis and machine intelligence}, 2026.

\bibitem[Qing et~al.(2023)Qing, Cai, Yang, and Yang]{qing2023story}
Zhongfei Qing, Zhongang Cai, Zhitao Yang, and Lei Yang.
\newblock Story-to-motion: Synthesizing infinite and controllable character animation from long text.
\newblock In \emph{SIGGRAPH Asia 2023 technical communications}, pages 1--4. 2023.

\bibitem[Sander et~al.(2022)Sander, Ablin, and Peyr{\'e}]{sander2022residual}
Michael Sander, Pierre Ablin, and Gabriel Peyr{\'e}.
\newblock Do residual neural networks discretize neural ordinary differential equations?
\newblock \emph{Advances in Neural Information Processing Systems}, 35:\penalty0 36520--36532, 2022.

\bibitem[Sui et~al.(2026)Sui, Ghosh, Hwang, Zhou, Wang, and Guo]{sui2026survey}
Kewei Sui, Anindita Ghosh, Inwoo Hwang, Bing Zhou, Jian Wang, and Chuan Guo.
\newblock A survey on human interaction motion generation.
\newblock \emph{International Journal of Computer Vision}, 134\penalty0 (3):\penalty0 113, 2026.

\bibitem[Tevet et~al.(2022)Tevet, Raab, Gordon, Shafir, Cohen-Or, and Bermano]{tevet2022human}
Guy Tevet, Sigal Raab, Brian Gordon, Yonatan Shafir, Daniel Cohen-Or, and Amit~H Bermano.
\newblock Human motion diffusion model.
\newblock \emph{arXiv preprint arXiv:2209.14916}, 2022.

\bibitem[Tong et~al.(2025)Tong, Nguyen-Tang, Lee, Nguyen, Tran, Hall, Kang, and Choi]{tong2025neural}
Anh Tong, Thanh Nguyen-Tang, Dongeun Lee, Duc Nguyen, Toan Tran, David Hall, Cheongwoong Kang, and Jaesik Choi.
\newblock Neural ode transformers: Analyzing internal dynamics and adaptive fine-tuning.
\newblock In \emph{International Conference on Learning Representations}, pages 52653--52686, 2025.

\bibitem[Xie et~al.(2026)Xie, Zheng, Han, Shi, Zhang, Bai, and Li]{xie2026textop}
Weiji Xie, Jiakun Zheng, Jinrui Han, Jiyuan Shi, Weinan Zhang, Chenjia Bai, and Xuelong Li.
\newblock Textop: Real-time interactive text-driven humanoid robot motion generation and control.
\newblock \emph{arXiv preprint arXiv:2602.07439}, 2026.

\bibitem[Zhang et~al.(2023{\natexlab{a}})Zhang, Zhang, Cun, Zhang, Zhao, Lu, Shen, and Shan]{zhang2023generating}
Jianrong Zhang, Yangsong Zhang, Xiaodong Cun, Yong Zhang, Hongwei Zhao, Hongtao Lu, Xi Shen, and Ying Shan.
\newblock Generating human motion from textual descriptions with discrete representations.
\newblock In \emph{Proceedings of the IEEE/CVF conference on computer vision and pattern recognition}, pages 14730--14740, 2023{\natexlab{a}}.

\bibitem[Zhang et~al.(2025)Zhang, Fan, and Yang]{zhang2025energymogen}
Jianrong Zhang, Hehe Fan, and Yi Yang.
\newblock Energymogen: Compositional human motion generation with energy-based diffusion model in latent space.
\newblock In \emph{2025 IEEE/CVF Conference on Computer Vision and Pattern Recognition (CVPR)}, pages 17592--17602. IEEE, 2025.

\bibitem[Zhang et~al.(2023{\natexlab{b}})Zhang, Guo, Pan, Cai, Hong, Li, Yang, and Liu]{zhang2023remodiffuse}
Mingyuan Zhang, Xinying Guo, Liang Pan, Zhongang Cai, Fangzhou Hong, Huirong Li, Lei Yang, and Ziwei Liu.
\newblock Remodiffuse: Retrieval-augmented motion diffusion model.
\newblock In \emph{2023 IEEE/CVF International Conference on Computer Vision (ICCV)}, pages 364--373. IEEE, 2023{\natexlab{b}}.

\bibitem[Zhang et~al.(2024{\natexlab{a}})Zhang, Cai, Pan, Hong, Guo, Yang, and Liu]{zhang2024motiondiffuse}
Mingyuan Zhang, Zhongang Cai, Liang Pan, Fangzhou Hong, Xinying Guo, Lei Yang, and Ziwei Liu.
\newblock Motiondiffuse: Text-driven human motion generation with diffusion model.
\newblock \emph{IEEE transactions on pattern analysis and machine intelligence}, 46\penalty0 (6):\penalty0 4115--4128, 2024{\natexlab{a}}.

\bibitem[Zhang et~al.(2024{\natexlab{b}})Zhang, Liu, Chen, Chen, Reid, Hartley, Zhuang, and Tang]{zhang2024infinimotion}
Zeyu Zhang, Akide Liu, Qi Chen, Feng Chen, Ian Reid, Richard Hartley, Bohan Zhuang, and Hao Tang.
\newblock Infinimotion: Mamba boosts memory in transformer for arbitrary long motion generation.
\newblock \emph{arXiv preprint arXiv:2407.10061}, 2024{\natexlab{b}}.

\bibitem[Zheng et~al.(2026)Zheng, Jin, Liu, and Zhao]{zheng2026next}
Zhiwei Zheng, Shibo Jin, Lingjie Liu, and Mingmin Zhao.
\newblock Next-scale autoregressive models for text-to-motion generation.
\newblock \emph{arXiv preprint arXiv:2604.03799}, 2026.

\bibitem[Zhong et~al.(2023)Zhong, Hu, Zhang, and Xia]{zhong2023attt2m}
Chongyang Zhong, Lei Hu, Zihao Zhang, and Shihong Xia.
\newblock Attt2m: Text-driven human motion generation with multi-perspective attention mechanism.
\newblock In \emph{2023 IEEE/CVF International Conference on Computer Vision (ICCV)}, pages 509--519. IEEE, 2023.

\bibitem[Zou et~al.(2024)Zou, Yuan, Du, Wang, Liu, Xu, Chen, and Ji]{zou2024parco}
Qiran Zou, Shangyuan Yuan, Shian Du, Yu Wang, Chang Liu, Yi Xu, Jie Chen, and Xiangyang Ji.
\newblock Parco: Part-coordinating text-to-motion synthesis.
\newblock In \emph{European conference on computer vision}, pages 126--143. Springer, 2024.

\end{thebibliography}
}


\end{document}